\pdfoutput=1

\documentclass[11pt]{article}
\usepackage[whole]{bxcjkjatype}

\usepackage[final]{acl}

\usepackage{float}
\usepackage{amsmath}
\usepackage{comment}
\usepackage{times}
\usepackage{here}
\usepackage{latexsym}
\usepackage[T1]{fontenc}
\usepackage[utf8]{inputenc}

\usepackage{inconsolata}
\usepackage{booktabs} 
\usepackage{graphicx}
\usepackage{multirow}
\usepackage{subcaption}
\usepackage{hhline}
\usepackage{url}
\title{Exploring the Relationship Between Diversity and Quality\\in Ad Text Generation}

\author{Yoichi Aoki${}^{1,2}$,  
        Soichiro Murakami${}^{3}$,
        Ukyo Honda${}^{3}$,
        Akihiko Kato${}^{3}$ \\
         ${}^{1}$Tohoku University,
         ${}^{2}$RIKEN, 
         ${}^{3}$CyberAgent \\ 
        \texttt{youichi.aoki.p2@dc.tohoku.ac.jp, } \\
        \texttt{\{murakami\_soichiro,honda\_ukyo,kato\_akihiko\}@cyberagent.co.jp, } \\
        }

\begin{document}
\maketitle
\begin{abstract}
In natural language generation for advertising, creating diverse and engaging ad texts is crucial for capturing a broad audience and avoiding advertising fatigue. Regardless of the importance of diversity, the impact of the diversity-enhancing methods in ad text generation---mainly tested on tasks such as summarization and machine translation---has not been thoroughly explored. Ad text generation significantly differs from these tasks owing to the text style and requirements. This research explores the relationship between diversity and ad quality in ad text generation by considering multiple factors, such as diversity-enhancing methods, their hyperparameters, input--output formats, and the models.
\end{abstract}

\section{Introduction}
Advertising is crucial for companies to promote their products and services to a broad audience. Research on natural language generation for advertising has advanced using language models to meet the demand for automating the ad creation ~\cite{HughesCZ19,KamigaitoZTO21,GolobokovCDGCCYL22,murakami2023}.

Diversity is a key metric in ad text generation. Repeatedly displaying the same ad to users can lead to boredom or advertising fatigue~\cite{pechmanAdvertisingRepetitionCritical1988,schmidt_advertising_2015}. Additionally, creating diverse ad texts will likely appeal to a broad range of customers. Therefore, a need exists for technology capable of generating diverse ad texts.

\begin{figure}[t]
  \includegraphics[width=0.9\columnwidth]{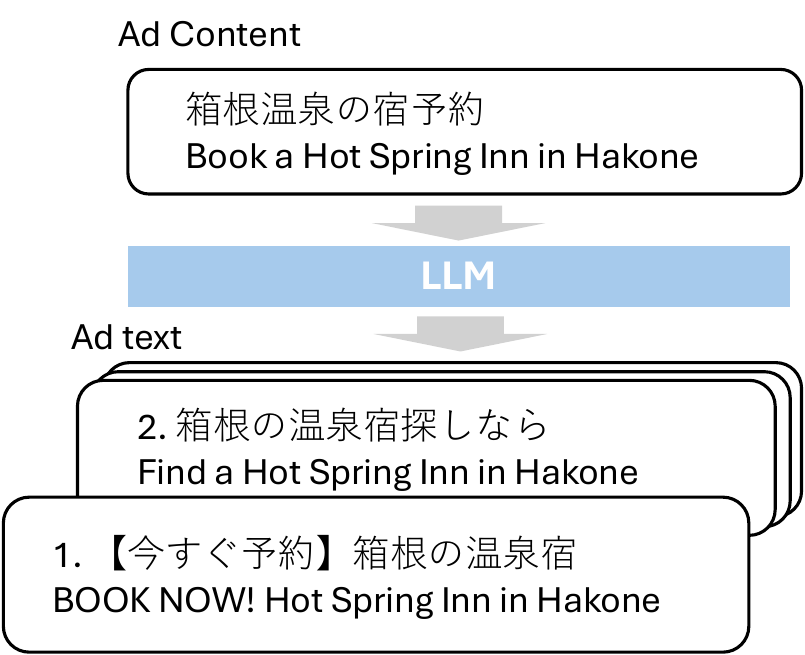}
  \caption{We input ad content into large language models to generate multiple ad texts. We measure the diversity and quality of the generated ads and elucidate the relationship between these aspects.}
  \label{figure:figure1}
\end{figure}
However, the impact of enhancing diversity on ad quality and other metrics is poorly understood when diversity-enhancing methods such as beam search~\cite{Bruce1976,Graves2012,RushCC13} or sampling~\cite{AckleyHS85,LewisDF18,HoltzmanBDFC20} in ad text generation with language models. Previous studies demonstrated that the relationship between diversity and quality varies significantly depending on the task~\cite{MeisterWC22}. While tasks such as summarization and machine translation (MT) exhibit a trade-off, other such as story generation exhibit interdependencies. Dialog tasks exhibit intermediate trends. In contrast, ad text generation differs from these tasks in the following ways:
\begin{enumerate}
    \item Measures such as ad performance and text length constraints are emphasized while minor grammatical errors are tolerable~\cite{Peinan2024}.
    \item Owing to the necessity of conveying messages effectively in limited spaces (including slogans or banner ads), unique expressions involving symbols or keywords are prevalent~\cite{murakami-etal-2025-adparaphrase}.
\end{enumerate}
The relationship between diversity and ad quality is non-trivial and requires systematic investigation, owing to these evaluation criteria and stylistic differences.

This study reveals that enhancing diversity involves trade-offs across multiple ad quality measures, and that sampling and beam search behave differently depending on the number of few-shot examples and outputs. These findings highlight the need for the careful use of diversity-enhancing methods in ad text generation. Moreover, we also provide promising directions for enhancing diversity while maintaining ad quality by combining the outputs of a set of different models.

\section{Problem Setting}
\subsection{Input and Output}
\label{subsec_input_output}
Following~\citet{murakami-etal-2025-adparaphrase}, we input the Japanese ad text dataset CAMERA~\cite{MitaMKZ24} as the ad content to a large language model and generated five diverse ad texts using diversity-enhancing methods (Figure~\ref{figure:figure1}). For in-context learning, we supplied three input--output examples to the model.

\subsection{Diversity-Enhancing Methods}
\label{subsec_diversity_enhancing_methods}
The primary decoding methods for achieving diversity include sampling and beam search~\cite{ZarriessVS21}. In this study, we use nucleus sampling~\cite{HoltzmanBDFC20}, temperature sampling~\cite{AckleyHS85}, beam search~\cite{Bruce1976,Graves2012,RushCC13}, and diverse beam search~\cite{LiMJ16} as typical sampling and beam search methods for generating diverse ad texts. We also employ Diverse MBR (DMBR) and k-medoids MBR (KMBR)~\cite{JinnaiHMZ24}, which offer better trade-offs in tasks such as translation and captioning.

\subsection{Evaluation}
\label{subsec_evaluation}
We measure the diversity among the five generated ad texts and the quality of each ad text.
\subsubsection{Ad Diversity}
Common contexts for diversity have two main aspects: \textbf{surface diversity (how to say?)}, and \textbf{semantic diversity (what to say?)}~\cite{murakami2023}.
In advertising, product details must be fixed. Therefore, we do not consider semantic diversity and \textbf{evaluat only the surface diversity among the outputs} in this study. We use the pairwise-BLEU similarity measure~\cite{PapineniRWZ02}, ranging from $0$ to $1$, based on n-gram matching, to evaluate the output diversity. Specifically, the diversity is calculated as $1-$Pairwise-BLEU.

\subsubsection{Ad Quality}
We evaluate ad quality based on three representative metrics defined by~\cite{Peinan2024}: ad performance, consistency, and acceptability. The average quality of the five generated outputs is considered the ad quality for one input.

\paragraph{Ad Performance:} Following~\citet{MitaMKZ24}, this study simulates customer behavior through click-through rates (CTR) from past distribution history to measure ad performance.\footnote{We utilize 極予測TD (Kiwami Yosoku TD), the CTR prediction model. The prediction aligns with CTR. cf. \url{https://cyberagent.ai/products/}} In the experiments, we report the performance ratio of the generated ads to reference human-written ads (generated/referenced) included in the CAMERA dataset.

\paragraph{Ad Consistency:} Generating ad texts with different ad content from the input potentially damages advertisers. For instance, converting the information "paid" to "free" can be false advertising. We evaluate the consistency between the ad content and the generated ad text using BERTScore~\cite{ZhangKWWA20}.

\paragraph{Ad Acceptability:} Ad platforms often impose length restrictions. We determine acceptability based on text fitting within 15 full-width or 30 half-width characters.\footnote{These are general restrictions in Japanese ad text on ad platforms such as Google Ads. cf. \url{https://support.google.com/google-ads/answer/1704389?hl=ja}}

\begin{figure*}[t]
  \includegraphics[width=2\columnwidth]{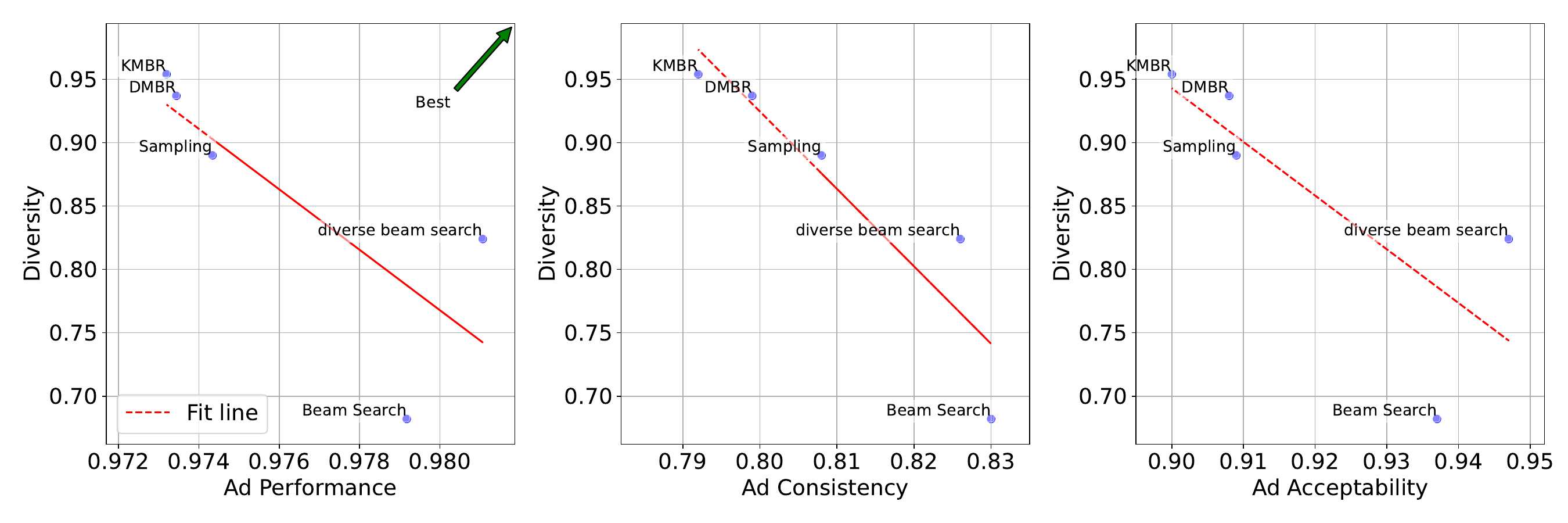}
  \caption{This figure shows the relationship between diversity and ad quality in ad text generation. Ad performance is the ratio of the performance of the generated ad text to the reference human-written ad text (generated/reference). Ad consistency is the BERT Score, whereas Ad acceptability is the percentage that meets the sentence length constraint. The upper right of the figure indicates higher diversity and ad quality.}
  \label{figure:diversity_enhancing_methods_results}
\end{figure*}
\section{Experiment}
In this experiment, we used calm3-22b-chat~\cite[calm3;][]{cyberagent-calm3-22b-chat}, Llama-3-ELYZA-JP-8B~\cite[ELYZA;][]{elyzallama2024}, Mistral-Small-24B-Instruct-2501~\cite[Mistral;][]{mistral_small_3}, Llama-3.1-Swallow-70B-Instruct-v0.3~\cite[Swallow;][]{Fujii:COLM2024,Okazaki:COLM2024,ma:arxiv2025}, GPT-4o~\cite{abs-2410-21276}. Hereinafter, we described only calm3's results and those of the other models in the Appendix ~\ref{section:appendix_other_models_results}.

\subsection{Diversity vs. Ad Quality Across~Diversity-Enhancing~Methods}
\label{subsection:diversity_enhancing_methods_experiment}
As shown in Figure~\ref{figure:diversity_enhancing_methods_results}, we found a trade-off between diversity and ad quality, with each value representing the average across the dataset. This suggests that the nature of ad generation differs from that of tasks such as dialogue and story generation~\cite{MeisterWC22}. This difference may be due to a more creative and dynamic dialogue and story generation. Compared with beam search, diverse beam search improved diversity while maintaining advertising performance and acceptability. Similar results were observed when the prompts or models were varied (see Appendices ~\ref{section:appendix_other_models_results} and~\ref{section:appendix_changing_prompt_results}). An exception was GPT-4o, which balanced improvements in diversity and acceptability. The human evaluation results were consistent with these results (see Appendix~\ref{section:appendix_human_evaluation}).

\subsection{Changing Parameters of Diversity-Enhancing Methods and Their Effects on Diversity and Ad Quality}
The relationship between diversity and ad quality when varying parameters for typical diversity-enhancing methods, such as sampling and beam search, is summarized in Table~\ref{table:parameter_results}. The sampling exhibited a trade-off. Beam search improved the performance and consistency up to a certain beam width, unlike previous studies on MT~\cite{KoehnK17}. This may be because ad generation produces shorter text than MT. On the other hand, beyond five beam widths, the beam width of the diverse beam search did not enhance the diverse beam search results.

\subsection{Effect of Number of Shots on~Diversity~and~Ad~Quality}
Table~\ref{table:few-shot_result} lists the effects of changing the number of few-shot examples. As diverse beam search re-
\begin{table}[H]
\centering
\begin{tabular}{|c|c|c|c|c|} 
\hline
\multicolumn{5}{|c|}{\textbf{sampling}} \\ \hline
param. & Div. & Per. & Con. & Accept. \\ \hline
P=0.5,T=1 &0.515  &\textbf{0.981} & \textbf{0.832}& \textbf{0.940} \\ 
P=1,T=1 & 0.890&0.974& 0.808&0.909  \\ 
P=1,T=1.5 &0.967 &  0.967& 0.773& 0.809 \\ 
P=1,T=2 & \textbf{0.991}&0.963 & 0.740& 0.619 \\ \hhline{=====}
\multicolumn{5}{|c|}{\textbf{beam search}} \\ \hline
param. & Div. & Per. & Con. & Accept. \\ \hline
W = 5 &0.682 &0.979  & 0.803&  0.937\\ 
W = 7 &0.697& \textbf{0.980}& \textbf{0.832} & \textbf{0.939}  \\ 
W = 10 & \textbf{0.702} & \textbf{0.980} & 0.831&  0.932\\ 
W = 12 &0.699 & \textbf{0.980} & \textbf{0.832} & 0.933 \\ \hhline{=====}
\multicolumn{5}{|c|}{\textbf{diverse beam search}} \\ \hline
param. & Div. & Per. & Con. & Accept. \\ \hline
W,G = 5 & \textbf{0.824}& \textbf{0.981} &0.826 & \textbf{0.947} \\ 
W,G = 7 & 0.814& 0.980& 0.825&0.941  \\ 
W,G = 10 &0.762 & 0.980 &0.826&  0.927\\ 
W,G = 12 &0.739 & 0.980 & \textbf{0.827} & 0.913 \\ \hline
\end{tabular}
\caption{The diversity and ad quality when varying the parameter of diversity-enhancing methods. Diversity, Ad Performance, Ad Consistency, and Ad Acceptability are denoted as Div., Per., Con., and Accept., respectively. The same notation is used in the following tables. 
We varied the p-value (P) and temperature (T) for sampling, the beam width (W) for beam search, and both the beam width (W) and group number (G) for diverse beam search.}
\label{table:parameter_results}
\end{table}
\noindent
sults were akin to a beam search, we described the results in Appendix ~\ref{section:appendix_diverse_beam_search_results}.
In sampling, diversity did not improve with increasing the number of shots. In contrast, in beam search and diverse beam search, diversity improved as the number of shots increased. This result is likely because seeing diverse examples led to changes such that the vocabulary with high output probability becomes more diverse.
\begin{table}[t]
\centering
\begin{tabular}{|c|c|c|c|c|}
\hline
\multicolumn{5}{|c|}{\textbf{sampling}} \\ \hline
shot num & Div. & Per. & Con. & Accept. \\ \hline
3  & \textbf{0.890} & 0.974 & 0.808 & 0.909 \\
9 & 0.872 & 0.985 &  0.822& 0.975 \\
15 & 0.876 & \textbf{0.987} & \textbf{0.825} & \textbf{0.982} \\\hline
\end{tabular}
\begin{tabular}{|c|c|c|c|c|}
\hline
\multicolumn{5}{|c|}{\textbf{beam search}} \\ \hline
shot num & Div. & Per. & Con. & Accept. \\ \hline
3 & 0.682 & 0.979 & 0.830 & 0.937 \\
9 & 0.738 &0.989  &0.845  & 0.982 \\
15 & \textbf{0.747} & \textbf{0.992}  & \textbf{0.849} & \textbf{0.987} \\\hline
\end{tabular}
\caption{Diversity and ad quality when varying the number of shots}
\label{table:few-shot_result}
\end{table}

\subsection{Effects of Output Number on~Diversity~and~Ad~Quality}
\label{subsection:output_num_experiment}
We investigated the diversity and ad quality when varying the number of outputs from five. The results are listed in Table~\ref{table:output-num_result}. Contrary to our intuition, consistent diversity and ad quality were maintained in the sampling, irrespective of the output number. In contrast, beam search improved diversity with more output but diminished quality, likely owing to the low quality of low probability candidate texts. Diverse beam search exhibited trends similar to those of the beam search (§~\ref{section:appendix_diverse_beam_search_results}).
\begin{table}[t]
\centering
\begin{tabular}{|c|c|c|c|c|}
\hline
\multicolumn{5}{|c|}{\textbf{sampling}} \\ \hline
output num & Div. & Per. & Con. & Accept. \\ \hline
2 & \textbf{0.890}&\textbf{0.975}  &0.807 &0.900  \\
5 & \textbf{0.890}&0.974  &\textbf{0.808} &\textbf{0.909} \\
10 & \textbf{0.890}&\textbf{0.975}  &0.807 &0.902  \\\hline
\end{tabular}
\begin{tabular}{|c|c|c|c|c|}
\hline
\multicolumn{5}{|c|}{\textbf{beam search}} \\ \hline
output num & Div. & Per. & Con. & Accept. \\ \hline
2 & 0.630&\textbf{0.981}  &\textbf{0.836} &\textbf{0.937}  \\
5 & 0.699&0.900  &0.832 &0.933  \\
10 & \textbf{0.737}&0.979  &0.829 &0.925  \\\hline
\end{tabular}
\caption{Diversity and ad quality when varying the output number}
\label{table:output-num_result}
\end{table}
\subsection{Effect of Changing Output Strategy}
\label{subsection:output_strategy_experiment}
The settings for generating multiple texts are not only not limited to the repeated inference approach illustrated in Figure~\ref{figure:figure1}, but multiple sentences can also be generated consecutively within a single inference step. We call the former one-at-once and the latter all-at-once, respectively. Table~\ref{table:all_at_once_prompt} shows the actual prompt used in this method. Beam search and other methods are strictly different techniques when their output strategies are different. Therefore, only the sampling results for the two output strategies are summarized in Table~\ref{table:output_strategy_results}. all-at-once exhibited improved diversity but worsened quality compared with one-at-once, indicating a trade-off. This diversity improvement could be because all-at-once can output ad text by reviewing previous outputs.
\begin{table}[t]
\centering
\begin{tabular}{c|c|c|c|c}
strategy & Div. & Per. & Con. & Accept. \\ \hline
one-at-once & 0.876 & \textbf{0.987} & \textbf{0.825} & \textbf{0.982} \\
all-at-once & \textbf{0.900} & 0.984  & 0.821 & 0.968 \\
\end{tabular}
\caption{Diversity and ad quality of sampling in all-at-once and one-at-once settings.}
\label{table:output_strategy_results}
\end{table}

\subsection{Diverse Outputs via Multiple Models}
We compared generating five texts with a single model to generating one text with the five models described above (five models). The results are summarized in Table~\ref{table:multi_model_results}.
Although the ad quality of the five models averaged each model's performance, diversity reached its highest value. This result suggests that generation from multiple models is a strategy for producing more diverse advertising texts.
\begin{table}[t]
\centering
\begin{tabular}{c|c|c|c|c}
Model & Div. & Per. & Con. & Accept. \\ \hline
calm3 & 0.890 & 0.974&0.808 & 0.909\\
ELYZA &  0.877 &0.970 &0.781 & 0.864\\
Mistral & 0.886 & 0.984&\textbf{0.833} &0.877\\
Swallow & 0.868 & 0.962& 0.795& 0.503\\
GPT-4o& 0.777 &\textbf{0.990} & 0.823& \textbf{0.992}\\
\textbf{5 models} & \textbf{0.929} & 0.975& 0.808& 0.827\\
\end{tabular}
\caption{Diversity and ad quality with multi-model}
\label{table:multi_model_results}
\end{table}

\section{Conclusion}
In this study, we revealed that improving diversity often involves compromises across multiple ad quality metrics, and that sampling and beam search behaved differently depending on the number of few-shot examples and outputs. These findings highlight the need for the careful use of diversity-enhancing methods in ad text generation. Moreover, we also provide promising directions for enhancing diversity while maintaining ad quality by combining the outputs of a set of different models. We believe that these insights will support future research in advancing both the diversity and quality of ad text generation.

\section*{Limitations}
We do not cover all methods for changing output diversity. Nevertheless, widely used sampling, beam search, and state-of-the-art MBR decoding methods have been employed to provide valuable insights on the relationship between diversity and ad quality.

Our analyses are based on Japanese ad texts. Therefore, please note that some features such as character types are unique to the Japanese language, and we do not intend to apply our analysis results to all languages. However, we believe that other languages such as English and Chinese also have unique linguistic features. In this study, we focused on Japanese ad texts; however, we hope that this will pave the way for their development in different languages.

\section*{Ethics statement}
This study will not raise particular ethical concerns, considering that (i) no human experiments are conducted and (ii) our tasks do not involve ethically sensitive topics.

\section*{Acknowledgments}
This work was supported by JST SPRING Grant Number JPMJSP2114. This work was written independently, with minor phrasing assistance from a large language model (ChatGPT).

\bibliography{main}
\clearpage
\appendix
\section{Detailed Experiment Settings}
We used 798 ad texts from the CAMERA dataset, each containing up to 30 half-width characters or 15 full-width characters. We used the dataset under the terms of the Creative Commons Attribution-NonCommercial-ShareAlike 4.0, International License~\footnote{\url{https://creativecommons.org/licenses/by-nc-sa/4.0/}}. In our experiments, we utilized the open models (calm3, ELYZA, Mistral, and Swallow) from HuggingFace. As for the closed model (GPT-4o), we accessed it via the Python library for the OpenAI API. The model inputs are listed in Table~\ref{table:one_at_once_prompt}. The §~\ref{subsection:output_strategy_experiment} used the prompt in Table~\ref{table:all_at_once_prompt}.

The sampling methods performed inference five times and output one ad text each time. On the other hand, the beam search approach performed inference once and output the top five candidates obtained. The parameters for each diversity-enhancing method used in the experiment are listed in Table~\ref{table:setting}. Note that in §~\ref{subsection:output_num_experiment}, the beam search and diverse beam search had a fixed beam width of 15. We used NVIDIA A100 (80GB) GPUs for the generation. 

\begin{table}[H]
\centering
\begin{tabular}{c|c}
method & parameter \\ \hline
sampling & 
\begin{tabular}[c]{@{}c@{}}
top\_p=1.0, \\
temperture=1.0
\end{tabular} \\ \hline
DMBR & 
\begin{tabular}[c]{@{}c@{}}
div\_pen = 1.0
\end{tabular} \\ \hline
beam search & 
\begin{tabular}[c]{@{}c@{}}
num\_beams = 5
\end{tabular} \\ \hline
diverse beam search & 
\begin{tabular}[c]{@{}c@{}}
num\_beams = 5, \\
num\_beam\_groups = 5, \\
diversity\_penalty=1.0
\end{tabular}
\end{tabular}
\caption{Parameter of diversity-enhancing methods}
\label{table:setting}
\end{table}
\begingroup
\begin{table*}[t]
    \centering
    \small
    \begin{tabular}{p{0.9\linewidth}}
        \toprule
        You are a professional ad copywriter. You are responsible for creating search-linked ads. Please rephrase the provided ad copy according to the following conditions.\\\\
        \# Conditions\\
        - Write within 15 full-width characters.\\
        - Do not add new information to or remove existing information from the ad copy.\\
        - Below are examples of rephrasing. Please use these examples as a reference to diversify your rephrasing.\\
        \vspace{1mm}
        \textbf{Input:} Personalized tutoring for strong performance in junior high exams \#中学受験に強い個別指導塾\\
        \textbf{Output:} Aim for junior high success with personal tutoring \#個別塾で中学合格を目指す\\
        \vspace{1mm}
        \textbf{Input:} Earn high income with side jobs in Kichijoji \#高収入を得る吉祥寺の副業で稼ぐ\\
        \textbf{Output:} Make money with Kichijoji side jobs \#吉祥寺の副業でお金儲け\\
        \vspace{1mm}
        \textbf{Input:} Corporate-focused [Immediate Strength Recruitment] \#法人向け【即戦力採用】\\
        \textbf{Output:} Seeking immediate strength! Corporate recruitment \#即戦力求む！法人採用\\
        \vspace{1mm}
        \textbf{Input:} {Ad content}\\
        \bottomrule
    \end{tabular}
    \caption{Model input for ad text generation. For visibility, Japanese prompt is translated into English. On the right side of each input--output example, the actual input--output examples are shown. Subsequent prompts will also be presented in English.}
    \label{table:one_at_once_prompt}
\end{table*}
\endgroup
\begingroup
\begin{table*}[t]
    \centering
    \small
        \begin{tabular}{p{0.9\linewidth}}
            \toprule
            You are a professional copywriter specializing in search-linked advertisements. Paraphrase the provided ad text according to the following conditions.\\\\
            \# Conditions\\
            - Write within 15 full-width characters\\
            - Do not add new information or remove existing information from the ad text\\
            - Below are examples of paraphrasing. Use them as a reference to variably paraphrase the ad text.\\
            \vspace{1mm}
            \textbf{Input:} Personalized tutoring for strong performance in junior high exams \#中学受験に強い個別指導塾\\
            \textbf{Output1:} Aim for junior high success with personal tutoring \#個別塾で中学合格を目指す\\
            \textbf{Output2:} The road to junior high success, tutoring \# 中学合格の道、個別塾\\
            \textbf{Output3:} Tutoring for junior high exam preparation \# 個別塾、中学試験対策\\
            \textbf{Output4:} Personal tutoring for junior high success! \# 個別で中学合格へ！\\
            \textbf{Output5:} Personalized tutoring for junior high entrance exams \# 中学入試対策の個別塾\\
            \vspace{1mm}
            \textbf{Input:} Earn high income with side jobs in Kichijoji \#高収入を得る吉祥寺の副業で稼ぐ\\
            \textbf{Output1:} Make money with Kichijoji side jobs \#吉祥寺の副業でお金儲け\\
            \textbf{Output2:} Side income in Kichijoji \# 吉祥寺で副業収入\\\
            \textbf{Output3:} High-paying side job in Kichijoji \# 高収入の吉祥寺副業\\
            \textbf{Output4:} Earn in a side job in Kichijoji \# 副業で稼ぐ in 吉祥寺\\
            \textbf{Output5:} High-paying side job @ Kichijoji \# 高収入副業＠吉祥寺\\
            \vspace{1mm}
            \textbf{Input:} Corporate-focused [Immediate Strength Recruitment] \#法人向け【即戦力採用】\\
            \textbf{Output1:} Seeking immediate strength! Corporate recruitment \#即戦力求む！法人採用\\
            \textbf{Output2:} Immediate impact with corporate hiring \# 即戦力で法人採用\\
            \textbf{Output3:} Immediate impact personnel for corporate \# 法人向けの人材即戦力\\
            \textbf{Output4:} Corporate immediate impact hiring \# 法人用即戦力採用\\
            \textbf{Output5:} Immediate impact hiring for companies \# 企業向け即戦力採用\\
            \vspace{1mm}
            \textbf{Input:} {Ad content}\\
            \bottomrule
        \end{tabular}
        \caption{Model input for ad text generation in an all-at-once setting.}
    \label{table:all_at_once_prompt}
\end{table*}
\endgroup

\section{Results of Other Models}
\label{section:appendix_other_models_results}
\begin{figure*}[t]
  \includegraphics[width=2\columnwidth]{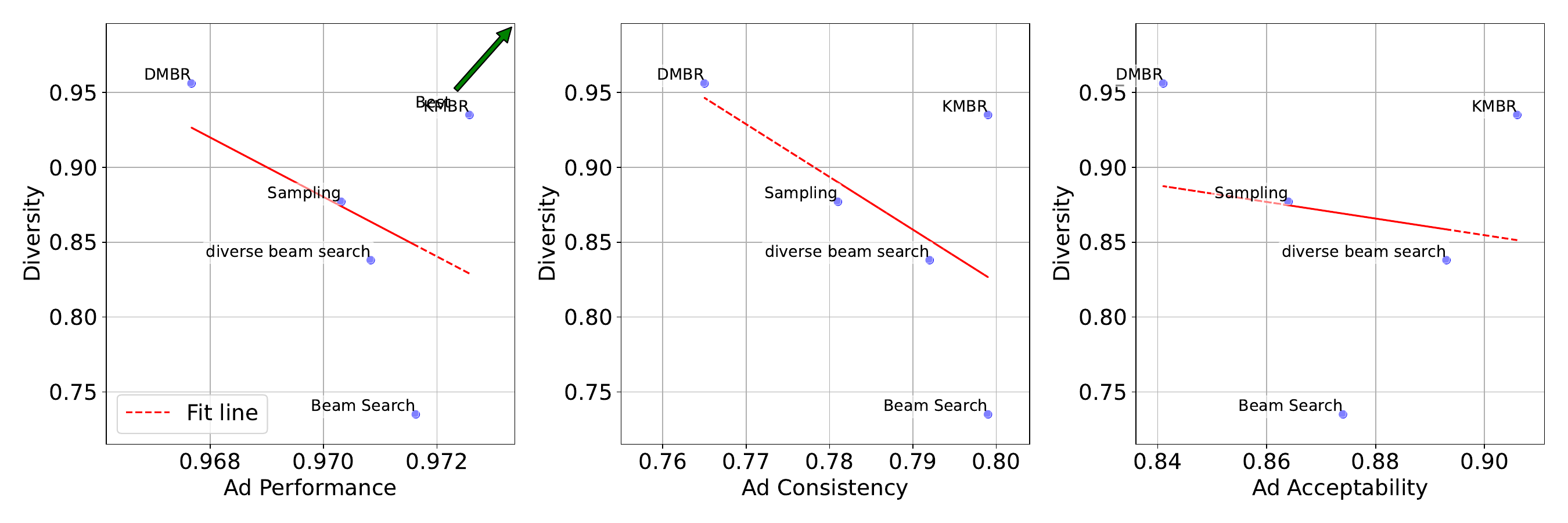}
  \caption{Relationship between diversity and ad quality in ad text generation with ELYZA}
  \label{figure:one_at_once_elyza}
\end{figure*}

\begin{figure*}[t]
  \includegraphics[width=2\columnwidth]{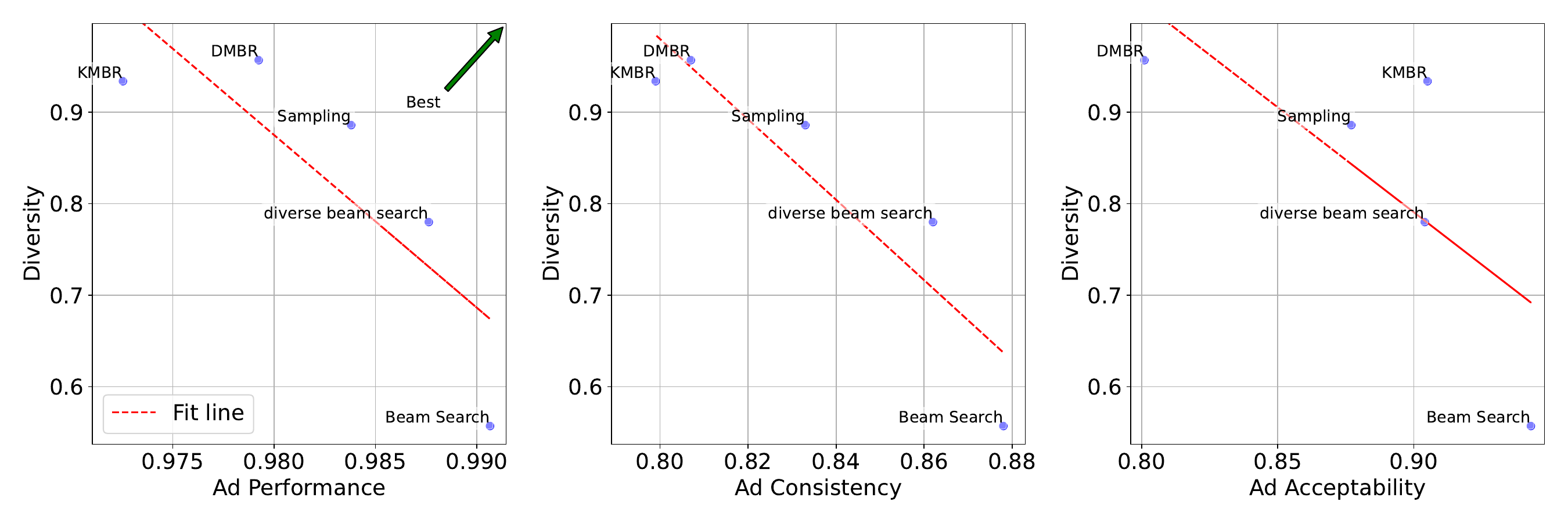}
  \caption{Relationship between diversity and ad quality in ad text generation with Mistral}
  \label{figure:one_at_once_mistral}
\end{figure*}

\begin{figure*}[t]
  \includegraphics[width=2\columnwidth]{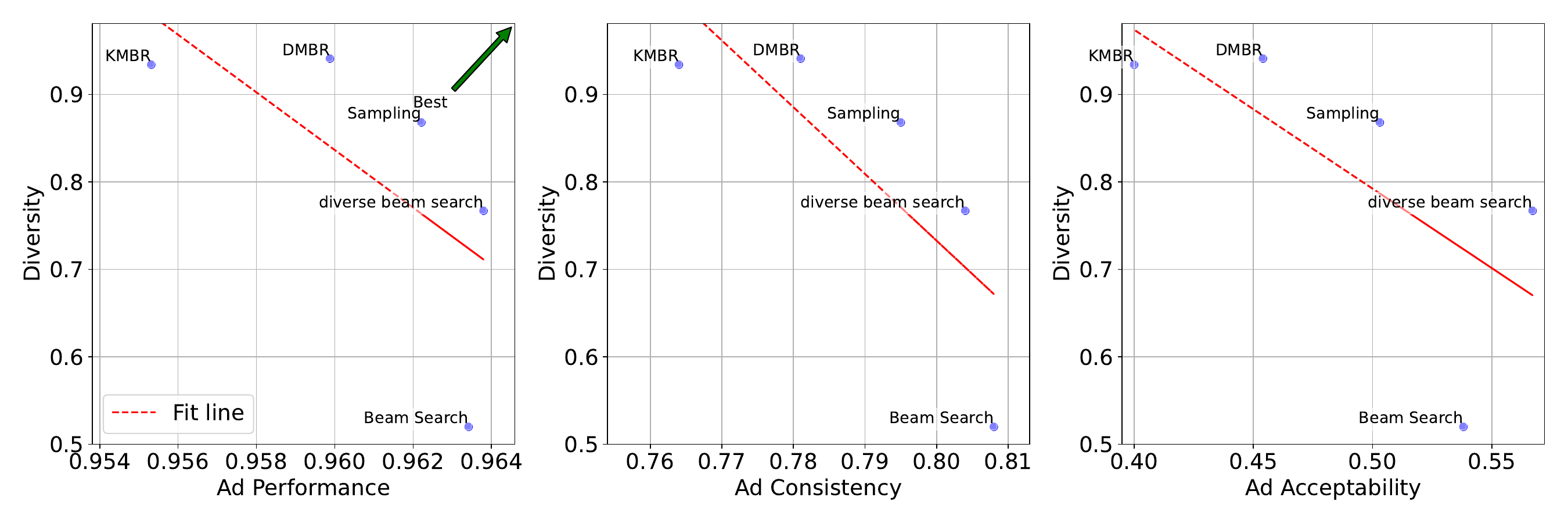}
  \caption{Relationship between diversity and ad quality in ad text generation with Swallow}
  \label{figure:one_at_once_swallow}
\end{figure*}

\begin{figure*}[t]
  \includegraphics[width=2\columnwidth]{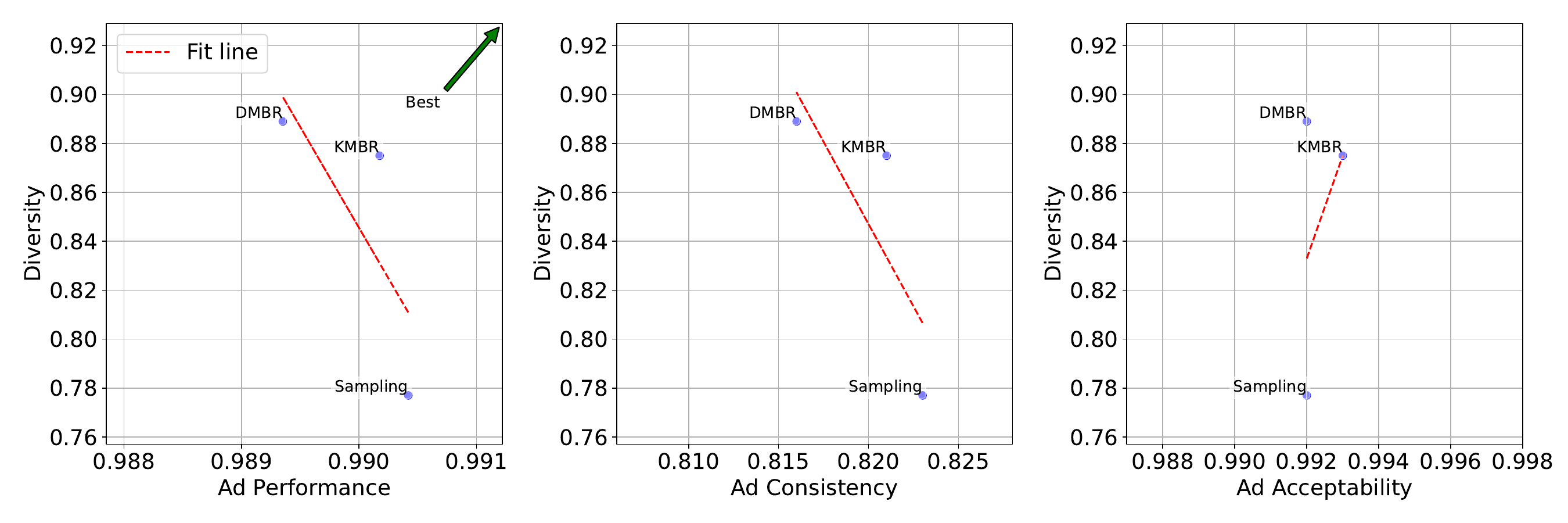}
  \caption{Relationship between diversity and ad quality in ad text generation with GPT-4o}
  \label{figure:one_at_once_gpt-4o}
\end{figure*}

The results for Llama-3-ELYZA-JP-8B (ELYZA), Mistral-Small-24B-Instruct-2501 (Mistral), Llama-3.1-Swallow-70B-Instruct-v0.3 (Swallow), and gpt-4o-2024-08-06 snapshot of OpenAI’s GPT-4o (GPT-4o) are shown in Figures~\ref{figure:one_at_once_elyza},~\ref{figure:one_at_once_mistral},~\ref{figure:one_at_once_swallow}, and ~\ref{figure:one_at_once_gpt-4o}, respectively. Consistently, we observed a trade-off between diversity and ad quality across multiple models. However, only GPT-4o balanced the enhancements in diversity and acceptability.

\section{Results Upon Changing the Prompt}
\label{section:appendix_changing_prompt_results}
\begin{figure*}[t]
  \includegraphics[width=2\columnwidth]{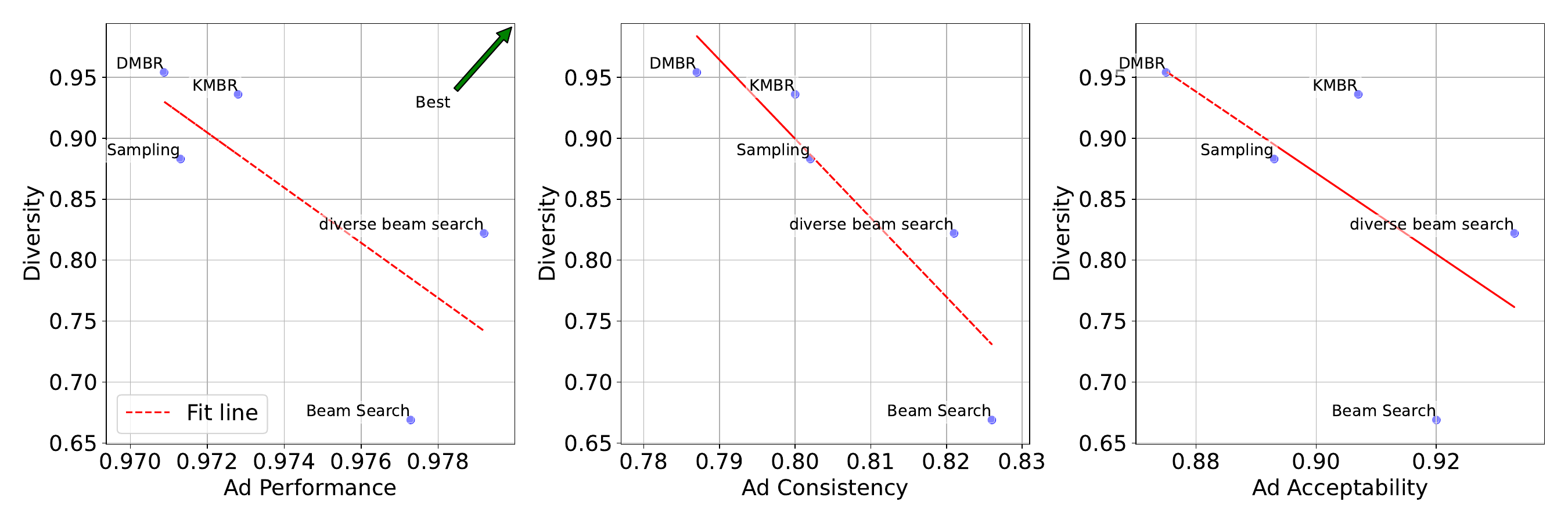}
  \caption{Diversity and ad quality of sampling when varying the order of few-shot examples}
  \label{figure:one_at_once_calm_prompt1}
\end{figure*}

\begin{figure*}[t]
  \includegraphics[width=2\columnwidth]{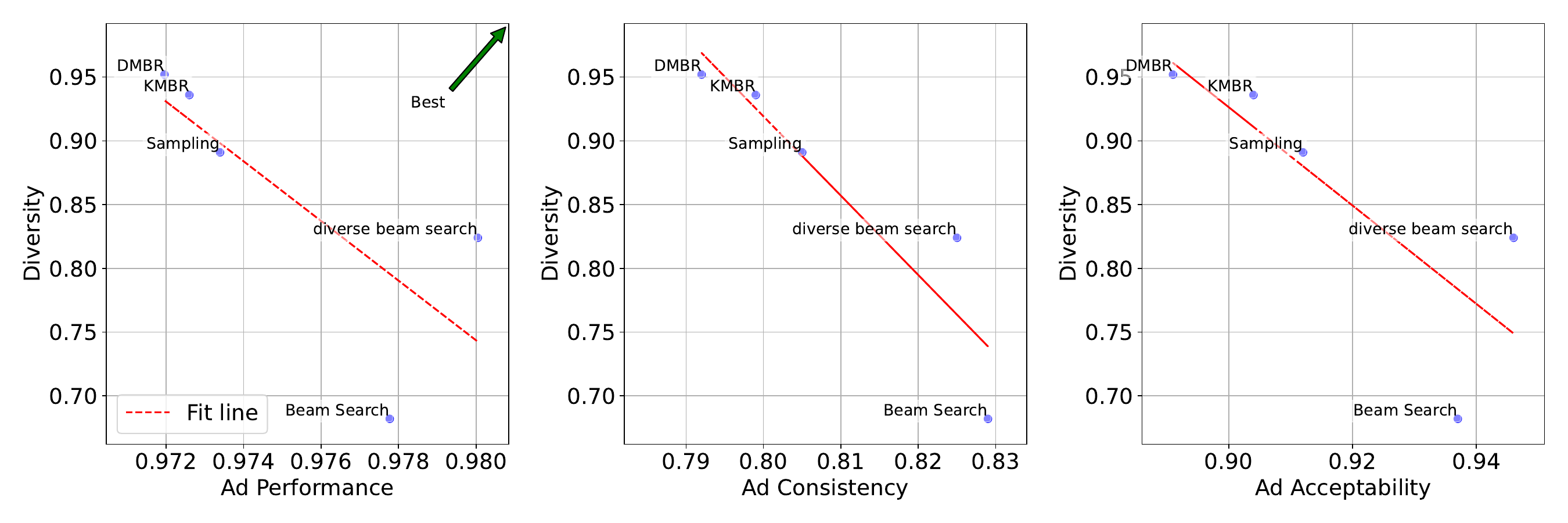}
  \caption{Diversity and ad quality of sampling when varying the instruction}
  \label{figure:one_at_once_calm_prompt2}
\end{figure*}

We examined the relationship between diversity and quality when varying the prompts, including (1) switching the order of few-shot examples and (2) altering explanations. The prompts for (2) are listed in Table~\ref{table:changed_prompt}. The results are shown in Figures~\ref{figure:one_at_once_calm_prompt1} and~\ref{figure:one_at_once_calm_prompt2}. Even when the prompts changed, it consistently demonstrated that a trade-off existed between diversity and ad quality.

\begingroup
\begin{table*}[t]
    \centering
    \small
    \begin{tabular}{p{0.9\linewidth}}
        \toprule
        You are an advertising expert. You have been tasked with creating listing ads. Based on the conditions below, rephrase the provided advertising text.\\\\
        \# Conditions\\
        - Keep it within 15 full-width characters\\
        - Do not add new information or remove included information from the ad text\\
        - Rephrase the ad text diversely using the examples below as references.\\
        \vspace{1mm}
        \textbf{Input:} Personalized tutoring for strong performance in junior high exams \#中学受験に強い個別指導塾\\
        \textbf{Output:} Aim for junior high success with personal tutoring \#個別塾で中学合格を目指す\\
        \vspace{1mm}
        \textbf{Input:} Earn high income with side jobs in Kichijoji \#高収入を得る吉祥寺の副業で稼ぐ\\
        \textbf{Output:} Make money with Kichijoji side jobs \#吉祥寺の副業でお金儲け\\
        \vspace{1mm}
        \textbf{Input:} Corporate-focused [Immediate Strength Recruitment] \#法人向け【即戦力採用】\\
        \textbf{Output:} Seeking immediate strength! Corporate recruitment \#即戦力求む！法人採用\\
        \vspace{1mm}
        \textbf{Input:} {Ad content}\\
        \bottomrule
    \end{tabular}
        \caption{Model input when varying the prompt.}
    \label{table:changed_prompt}
\end{table*}
\endgroup

\section{Consistency With Human and Automatic Evaluation}
\label{section:appendix_human_evaluation}
We evaluated the alignment between the automatic metrics and the human judgment used in our experiments. Five annotators performed a human evaluation of 1,000 generated advertising texts. For ad performance, the annotators selected a more attractive text from the reference and generated versions. Although the ad performance and attractiveness were not identical, the agreement rate between attractiveness as evaluated by humans and ad performance measured by automatic metrics was 61\%. For consistency, the annotators determined whether the generated text preserved the semantics of the input. Because consistency is not a relative evaluation, we report the Pearson correlation~\cite{Pearson1895} between the human and automatic scores, which was 0.55, indicating a moderate correlation.

The human evaluation was performed via Yahoo crowdsourcing, with a total compensation of 5,500 yen. The annotators received the instructions listed in Tables~\ref{table:human_instruction_ad_performance} and~\ref{table:human_instruction_ad_consistency}, and provided consent for the results of their evaluations to be used in this study.
\begingroup
\begin{table*}[t]
    \centering
    \small
    \begin{tabular}{p{0.9\linewidth}}
        \toprule
        Please select the more **attractive** ad text. If there is no difference in impression between the two ad texts, please click "No Difference".\\
        \vspace{1mm}
        【Evaluation Criteria】\\
        - Do you want to click on it?\\
        - Is it eye-catching?\\
        - Is it easy to read?\\
        \vspace{1mm}
        【Question】\\
        Please select the more **attractive** ad text.\\
        - \{Ad Text 1\}\\
        - \{Ad Text 2\}\\
        - No Difference\\
        \bottomrule
    \end{tabular}
    \caption{Instruction for human evaluation of attractiveness. For visibility, Japanese instruction is translated into English.}
    \label{table:human_instruction_ad_performance}
\end{table*}
\endgroup
\begingroup
\begin{table*}[t]
    \centering
    \small
    \begin{tabular}{p{0.9\linewidth}}
        \toprule
        【Question\\
        Please determine whether the information contained in the following two advertisements is **equivalent or not**.\\
        - \{Ad Text 1\}\\
        - \{Ad Text 2\}\\
        \vspace{1mm}
        【Evaluation Criteria】\\
        - Focus on differences in the information conveyed, not on variations in vocabulary or word order.\\
        - Even if certain information is not explicitly stated, information that can be inferred based on common sense is considered to be included in the ad text.\\
        \vspace{1mm}
        【Options】\\
        - Equivalent\\
        - Not Equivalent\\
        \bottomrule
    \end{tabular}
    \caption{Instruction for human evaluation of ad consistency. For visibility, Japanese instruction is translated into English.}
    \label{table:human_instruction_ad_consistency}
\end{table*}
\endgroup

\section{Changing Parameters of DMBR and~Their~Effects on Diversity and~Ad~Quality}
We showed the relationship between diversity and ad quality when varying parameters of DMBR in Table~\ref{table:parameter_dmbr_results}. Similar to sampling, DMBR exhibited a trade-off.
\begin{table}[H]
\centering
\begin{tabular}{c|c|c|c|c} 
param. & Div. & Per. & Con. & Accept. \\ \hline
D = 0.1 &  0.767& \textbf{0.978} & \textbf{0.826}& \textbf{0.922} \\ 
D = 0.5 &0.927 & 0.974 &0.802 &  0.904\\ 
D = 1&0.954 & 0.973 & 0.792& 0.9 \\ 
D = 2 & \textbf{0.957}& 0.973 &0.79 & 0.897\\ 
\end{tabular}
\caption{Diversity and ad quality of sampling when varying the parameter of DMBR. We vary diversity penalty (D) for DMBR.}
\label{table:parameter_dmbr_results}
\end{table}

\section{Results of Diverse Beam Search}
\label{section:appendix_diverse_beam_search_results}
The relationship between diversity and ad quality when varying few-shot numbers and output numbers are summarized in Tables~\ref{table:few-shot_diverse_beam_search_result} and~\ref{table:output-num_diverse_beam_search_result}. Overall results mirrored those of beam search. Increasing the number of shots impacted diversity increase. In contrast, increasing the number of outputs improved diversity but reduced quality.
\begin{table}[H]
\centering
\begin{tabular}{c|c|c|c|c}
shot num & Div. & Per. & Con. & Attempt. \\ \hline
3 & 0.824 &0.981  &0.826 &0.947  \\
9 & \textbf{0.832} &0.990&0.838&0.987\\
15 & 0.830&0.991&0.842& \textbf{0.991}\\
\end{tabular}
\caption{Diversity and ad quality of diverse beam search when varying the few-shot number}
\label{table:few-shot_diverse_beam_search_result}
\end{table}

\begin{table}[H]
\centering
\begin{tabular}{c|c|c|c|c}
output num & Div. & Per. & Con. & Accept. \\ \hline
2 & 0.781&\textbf{0.981}  &\textbf{0.829} &\textbf{0.932}  \\
5 & 0.736&0.979  &0.825 &0.905  \\
10 & 0.801&0.977  &0.819 &0.878  \\
\end{tabular}
\caption{Diversity and ad quality of diverse beam search when varying the output number}
\label{table:output-num_diverse_beam_search_result}
\end{table}

\section{Sampling in All-at-Once Setting}
This study also measured the diversity and ad quality of sampling by varying the parameters of the diversity-enhancing methods in an all-at-once setting. The results are presented in Figure~\ref{table:tradeoff_all_at_once_sampling}. The results when varying the few-shot numbers and parameters were similar in the all-at-once and one-at-once settings.
\begin{table}[H]
\centering
\begin{tabular}{c|c|c|c|c}
parameter & Div. & Per. & Con. & Accept. \\ \hline
P=0.5,T=1 &0.753 &\textbf{0.99}& \textbf{0.843}& \textbf{0.979} \\ 
P=1,T=1 &0.900 &0.984&0.821&0.968 \\ 
P=1,T=1.5 &0.960 & 0.976 &0.787 &0.908  \\ 
P=1,T=2 &\textbf{0.988} &0.969& 0.746& 0.724\\ 
\end{tabular}
\caption{Diversity and ad quality of sampling in an all-at-once setting when varying the parameter of diversity-enhancing methods.}
\label{table:tradeoff_all_at_once_sampling}
\end{table}

\section{Usage of AI assistants}
We used AI assistants (e.g., GPT4-o and GitHub Copilot) to write this paper and provide the source code for the experiments. However, its use was limited to code completion, translation, text editing, and table creation, and all the content was based solely on the authors' ideas.

\end{document}